\documentclass[letterpaper, 10 pt, conference]{ieeeconf}  

\IEEEoverridecommandlockouts                              

\usepackage{makeidx}  
\usepackage{graphicx}
\usepackage{cite}
\usepackage{cleveref}
\usepackage{amsmath}

\title{\LARGE \bf
ToolNet: Holistically-Nested Real-Time Segmentation of \\ 
Robotic Surgical Tools
}

\author{
	Luis C. Garc\'ia-Peraza-Herrera$^{1}$,
	Wenqi Li$^{1}$,
	Lucas Fidon$^{1}$,
	Caspar Gruijthuijsen$^{4}$,
	Alain Devreker$^{4}$, \\
	George Attilakos$^{3}$,
	Jan Deprest$^{1,3,5,6}$,
	Emmanuel Vander Poorten$^{4}$,
	Danail Stoyanov$^{2,6}$, \\
	Tom Vercauteren$^{1,6}$,
	S\'ebastien Ourselin$^{1,6}$
\thanks{$^{1}$Translational Imaging Group, CMIC, Dept. Med. Phys. Biomed. Eng., University College London, UK {\tt\small luis.herrera.14@ucl.ac.uk}}%
\thanks{$^{2}$Surgical Robot Vision Group, CMIC, Dept. Computer Science, University College London, UK}%
\thanks{$^{3}$University College London Hospitals, UK}%
\thanks{$^{4}$Katholieke Universiteit Leuven, Belgium}%
\thanks{$^{5}$Universitair Ziekenhuis Leuven, Belgium}%
\thanks{$^{6}$Wellcome / EPSRC Centre for Surgical and Interventional Sciences, UCL, London, UK}%
}

\begin{document}

\maketitle
\thispagestyle{empty}
\pagestyle{empty}

\begin{abstract}

Real-time tool segmentation from endoscopic videos is an essential part of many computer-assisted robotic surgical systems and of critical importance in robotic surgical data science. We propose two novel deep learning architectures for automatic segmentation of non-rigid surgical instruments. Both methods take advantage of automated deep-learning-based multi-scale feature extraction while trying to maintain an accurate segmentation quality at all resolutions. The two proposed methods encode the multi-scale constraint inside the network architecture. The first proposed architecture enforces it by cascaded aggregation of predictions and the second proposed network does it by means of a holistically-nested architecture where the loss at each scale is taken into account for the optimization process. As the proposed methods are for real-time semantic labeling, both present a reduced number of parameters. We propose the use of parametric rectified linear units for semantic labeling in these small architectures to increase the regularization of the network while maintaining the segmentation accuracy. We compare the proposed architectures against state-of-the-art fully convolutional networks. We validate our methods using existing benchmark datasets, including \textit{ex vivo} cases with phantom tissue and different robotic surgical instruments present in the scene. Our results show a statistically significant improved Dice Similarity Coefficient over previous instrument segmentation methods. We analyze our design choices and discuss the key drivers for improving accuracy.

\end{abstract}

\section{INTRODUCTION}

In recent years, the operating room (OR) has evolved towards a highly technological environment. Minimally invasive robotic surgery and computer-assisted surgical systems are now a clinical reality. Robotic tool detection, segmentation, tracking and pose estimation are bound to become core technologies with many potential applications in these complex surgical scenarios~\cite{Bouget2017}.

Minimally invasive robotic surgery increases the surgeon comfort while operating, facilitates control over anatomy and allows for dexterous manipulations within reduced operating spaces. However, the master-slave configuration of such systems induces a loss of direct contact with the tissue. Haptic cues are thus gone. Yet, they are a desirable feature which surgeons would like to resort to while operating. Haptic feedback helps to perform safe manipulations and confirm the correct execution of surgical treatments \cite{VanDerMeijden2009}. Tool segmentation represents an essential step towards providing improved haptics and virtual fixtures, increasing the context-awareness of surgeons whilst performing robotic interventions and potentially helping to reduce the number of complications in the OR \cite{Bouget2015}.

\begin{figure}[thpb]
	\centering
	\includegraphics[width=\columnwidth]{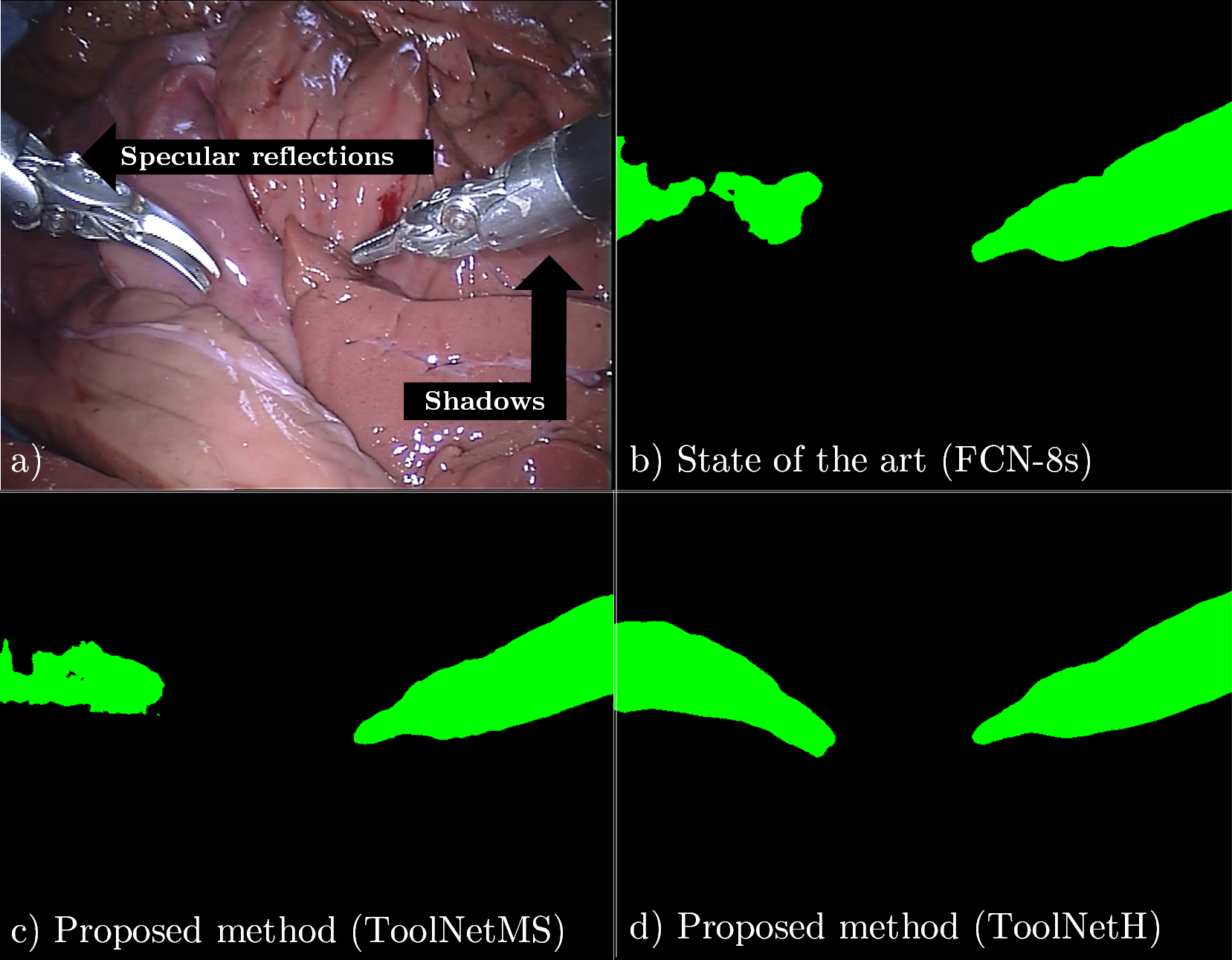}
	\caption{(a) Frame from the MICCAI 2015 Endoscopic Vision Challenge testing set \cite{MICCAIchallenge2015}. Challenges faced by instrument-tissue segmentation algorithms highlighted. (b) State-of-the-art FCN-8s segmentation prediction. (c) Segmentation output illustrating the proposed multi-scale method \texttt{ToolNetMS}. (d) Segmentation output illustrating the proposed holistically-nested method \texttt{ToolNetH}.}
	\label{fig:teaser}
\end{figure}

To facilitate minimally invasive surgery, robotic surgical instruments must be small and allow for highly dexterous manipulations. With the reduction in size, complex actuation systems are required. Long kinematic chains (12 active joints in the da Vinci robot~\cite{Freschi2013}, Intuitive Surgical, USA), micro-machined super-elastic tool guides with pneumatic artificial muscles~\cite{Devreker2015} and concentric tubes~\cite{Dwyer2017} are recent examples of these highly complex actuation mechanisms. As a consequence, the kinematics of such robotic manipulators become less stable due to hysteresis, friction and backlash. This makes the control of said instruments an extremely challenging task. Furthermore, as the dexterity of the instruments rises, it becomes increasingly difficult for the surgeon to both manipulate it and get a cognitive representation of the robot kinematics~\cite{Devreker2015}. To ease clinical translation, a high-level control should ideally be provided. Embedding shape-sensing or position-tracking technologies without increasing the instrument's size is technically very challenging. There are also surgical settings such as endoscopic fetal surgery in which this is not even possible, as increasing the size or the number of endoscopic ports is linked to adverse outcomes such as pre-term delivery and rupture of uterine membranes~\cite{Fowler2002}. Visual servoing represents a potential solution to control complex robotic manipulators. It can leverage the visualization capabilities that are already present in the OR. Precise real-time robotic tool segmentation is a building block that can be employed for this control strategy.



There are a number of challenges that turn segmentation of surgical tools into a remarkably difficult task (as illustrated in \cref{fig:teaser}). Endoscopic scenes are heterogeneous environments that feature a high variety of optical interactions that complicate the segmentation, namely: specular reflections, partial occlusions, blurriness (due to fast motion and/or parts of the image out of focus), reflections (from instrument onto the patient's tissue and vice versa), body fluids and smoke. A possible option to ease the vision-based segmentation is to use fiducial markers~\cite{Reiter2011}. However, this does not seem to be an viable solution. Coating surgical instruments alters the current surgical workflow as the vision system (e.g. real-time tracking) would be limited to work with customized tools only. This has a negative impact on the manufacturing process, possibly increasing the cost of the instruments, as the markers need to be able to withstand the sterilization process. Furthermore, all previously recorded \textit{surgical big data} could not easily be used, which would represent an important missed opportunity to robotic surgical data science~\cite{SwaroopVedula2016}.

\subsection{Related work}

Traditional methods that aim to separate surgical instruments from background tissue typically rely on hand-crafted feature extraction and selection. The input image is transformed into a higher dimensional feature space that takes into account color and texture. Then, a classifier is trained to discriminate between tools and tissue. However, due to the highly tangled appearance patterns of instruments and background in endoscopic images, manually designing discriminative features to separate the data becomes very challenging. Different machine learning techniques such as Boosted Decision Trees \cite{Bouget2015} and Random Forests \cite{Allan2013} have been employed for pixel-wise binary classification. Recently, EndoNet \cite{EndoNet} has successfully shown the ability of Convolutional Neural Networks (CNNs) to perform instrument detection and surgical phase recognition. DeepMedic \cite{Kamnitsas2016} has also demonstrated that Fully Convolutional Networks (FCNs) can perform well on segmentation tasks with medical images. In the context of instrument segmentation, Garcia-Peraza-Herrera et al. \cite{Peraza2016} recently  proposed a real-time tool segmentation pipeline that employs FCNs combined with optical flow tracking to reduce the computational burden by exploiting short-term temporal consistency.


\subsection{Contributions}

This paper presents, to the best of our knowledge, the first convolutional network architectures trained end-to-end for real-time semantic segmentation of robotic surgical tools. We introduce two novel lightweight architectures, \texttt{ToolNetMS} and \texttt{ToolNetH}. Both feature one order of magnitude fewer parameters than the state-of-the-art, requiring less memory and allowing for real-time inference.

We propose Dice loss as the objective function adapted to our problem and data. We also introduce the parametric rectified linear unit (PReLU)~\cite{He2015PRELU} as a suitable adaptively-learnt activation function for semantic labeling. This helps to improve segmentation accuracy at the expense of an insignificant additional computational cost.

Our architectures employ two novel ways of imposing multi-scale consistency within the predictions of the network. \texttt{ToolNetMS} aggregates multi-scale predictions and then calculates the final Dice loss.
\texttt{ToolNetH} introduces a new loss function that imposes multi-scale consistency by means of a holistically-nested approach. 

Our proposed algorithms outperforms the state-of-the-art baseline architecture on a challenging robotic instrument-background segmentation dataset with robotic instruments not present in training.

\section{METHODS}

\begin{figure*}[thpb]
	\centering
	\includegraphics[width=\textwidth]{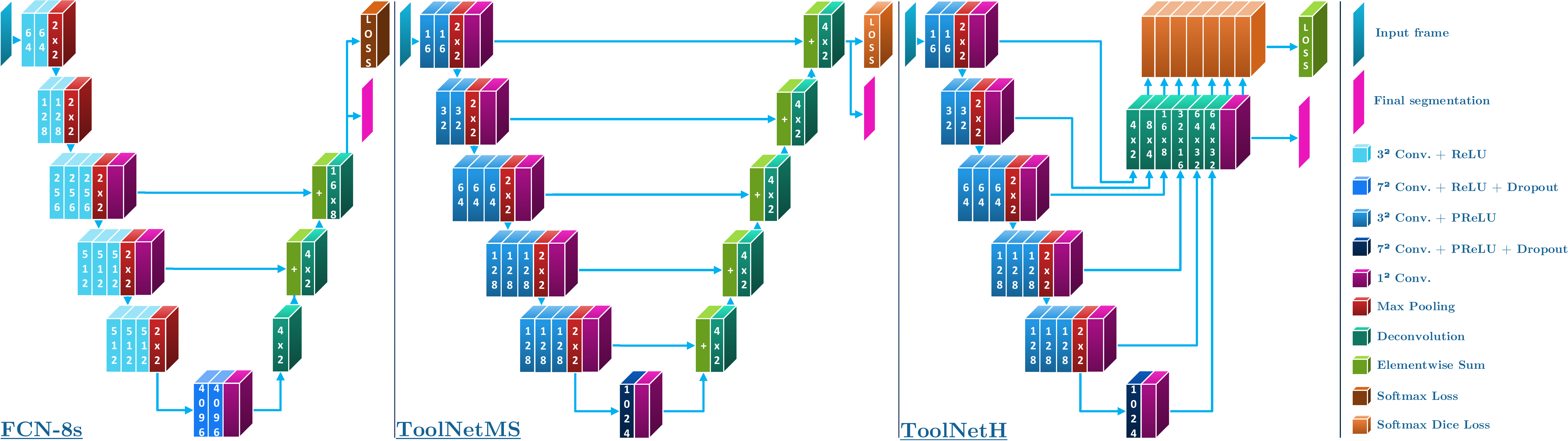}
	\caption{Proposed convolutional architectures (center and right) compared to state-of-the-art FCN-8s (left). Conv. stands for convolution. The number inside each layer indicates the amount of convolutional kernels.}
	\label{fig:cnn_architecture}
\end{figure*}

\subsection{Fully Convolutional Network (FCN) architectures}

Fully Convolutional Networks (FCNs) have been proposed \cite{Long2015} as a state-of-the-art machine learning approach to automatically extract discriminative pixel-level features from images and perform accurate pixel-wise segmentations. In contrast to previous convolutional neural networks (CNNs) such as AlexNet \cite{alexnet2012} or VGG16\cite{vgg16}, FCNs do not have fully connected layers. This makes them suitable for end-to-end pixel-level semantic labeling, as the spatial configuration of the image is preserved across the layers. 

FCNs have shown to perform well in general purpose segmentation tasks \cite{Zheng2015} as well as in instrument-background segmentation of surgical images \cite{Peraza2016}. Although on average they produce accurate pixel-wise label estimations, there are some limitations that need to be considered to make improvements over current results. FCNs generate the final pixel-wise labels by upsampling a prediction eight times smaller than the original image. Hence, the predicted masks often feature holes or do not respect edges. They also lack real-time capabilities and feature a loss function that is not adequate for unbalanced class data.



\subsection{Dice loss}

Cross-entropy loss is a stable error metric widely used in deep learning. However, in surgical scenes the number of pixels belonging to foreground and background differs by orders of magnitude. To improve the accuracy of our predictions we have to balance the contribution of overlap over classes. Furthermore, as we usually employ the Dice Similarity Coefficient (DSC) to evaluate the quality of the segmentations, it seems natural to incorporate it as our loss function to optimise during network training. This has been proposed by~\cite{Milletari2016a,Li2017} for segmentation of 3D volumes. In this work we propose to use it for 2D instrument-background segmentation, reformulating it as a minimization of the weighted mean squared error (MSE):
\begin{align}
%
%
%
%
\mathcal{L}_{\textrm{DSC}}(\mathbf{\hat{y}}, \mathbf{y}) =
\sum_{i=1}^{P}\sum_{k=1}^{K}\alpha (\mathbf{\hat{y}}, \mathbf{y}, k) (\hat{y}_{i,k} - y_{i,k})^2
%
\end{align}
\noindent where $P$ is the number of pixels in the image, $K$ is the number of classes (two in our case, instrument and background tissue), $\mathbf{\hat{y}}=\{\hat{y}_{i,k}\}_{1 \le i \le P,\, 1 \le k \le K}$ is the probabilistic label prediction, $\mathbf{y}=\{y_{i,k}\}_{1 \le i \le P,\, 1 \le k \le K}$ is the probabilistic ground truth label. $\alpha (\mathbf{\hat{y}}, \mathbf{y}, k)$ balances the estimation error by taking into account the amount of pixels belonging to each class so that errors in less represented classes such as the surgical instruments (which consist of a small amount of pixels) do not become negligible. $\forall k\in\{1, ..., K\}$, $\alpha (\mathbf{\hat{y}}, \mathbf{y}, k)$ is defined as:
\begin{align}
%
%
%
\alpha (\mathbf{\hat{y}}, \mathbf{y}, k) = \frac{1}{K} \frac{1}{\sum_{i=1}^{P} \hat{y}_{i,k}^2 + \sum_{i=1}^{P}y_{i,k}^2}
\end{align}

\subsection{Parametric Rectified Linear Unit (PReLU)}

As the network and training data get smaller compared to the \texttt{FCN-8s} (which can enjoy transfer learning from very deep classification networks such as VGG16), we require a more robust regularization strategy that maintains the generalization ability of the network and protects the architecture from overfitting the training set. An effective approach to achieve this is to make non-linear activation functions more robust. A recently proposed strategy in this direction is the Parametric Rectified Linear Unit (PReLU) \cite{He2015PRELU}. As opposed to the well-known rectified linear unit (ReLU), PReLU is a non-linear activation function which learns the negative slopes, providing end-to-end trained activations:
\begin{align}
f(x_c) = \max(0, x_c) + a_c \min(0, x_c)
\end{align}
\noindent where $x_c$ is the input of the non-linear activation $f$ on the $c^{th}$ channel. When $a_c$ is close to zero, $f(x_c)$ is equivalent to a Leaky ReLU \cite{Maas2013}, if it is exactly zero then $f(x_c)$ behaves as a conventional ReLU. As the weight on the negative slope is shared for each feature map, this adds an insignificant amount of extra parameters ($a_i$) while improving accuracy~\cite{He2015PRELU}. Hence, we have selected it as our activation function for all the layers that feature a rectified unit in the encoder part of the proposed networks.

\subsection{Proposed multi-scale and holistically-nested networks}

We propose two new architectures for real-time robotic instrument-background segmentation, \texttt{ToolNetMS} and \texttt{ToolNetH} (see \cref{fig:cnn_architecture}). Both make improvements in terms of accuracy and speed over the \texttt{FCN-8s}. 

\texttt{ToolNetMS} achieves more accurate segmentations around the tool-tissue boundaries because both the architecture and the loss function are tailored to the problem. By summing predictions at \textit{all} scales in a cascaded fashion we ensure that the responses around the edges are finer than in the \texttt{FCN-8s}. Dice loss is employed to take into account the class unbalance between instrument and tissue pixels. We also reduce the number of parameters compared to \texttt{FCN-8s} to be able to perform inference at the speed of robotic surgery video sequences (typically 25fps).

The second architecture that we propose and evaluate is \texttt{ToolNetH}. As opposed to \texttt{ToolNetMS}, this network does not have a cascaded element-wise sum between the class-scores provided at different scales. Instead, inspired by the holistically-nested edge detection\cite{Xie2015}, where the authors aggregate sigmoid cross-entropy losses from multiple scales, we propose a Multi-Scale Dice Loss (MSDL) for semantic segmentation that takes into account the accuracy of predictions at different scales:
\begin{align}
\mathbf{\hat{y}}^{(\bar{s})}(\mathbf{z}, \boldsymbol{\theta}) &= \sum_{j=1}^{M}w_j \mathbf{\hat{y}}^{(s_j)}(\mathbf{z}, \boldsymbol{\theta}) \\
\mathcal{L}_{\textrm{MSDL}}(\mathbf{y}, \mathbf{z}, \boldsymbol{\theta}) &=
\bar{\lambda}\mathcal{L}_{\textrm{DSC}}(\mathbf{\hat{y}}^{(\bar{s})}(\mathbf{z}, \boldsymbol{\theta}), \mathbf{y}) \\
\nonumber&+ \sum_{j=1}^{M}\lambda_j\mathcal{L}_{\textrm{DSC}}(\mathbf{\hat{y}}^{(s_j)}(\mathbf{z}, \boldsymbol{\theta}), \mathbf{y})
\end{align}
\noindent where $\mathbf{z}$ and $\boldsymbol{\theta}$ represent input image and weights of the network respectively. $\mathbf{\hat{y}}^{(s_j)}(\mathbf{z}, \boldsymbol{\theta})$ represents a probabilistic prediction at scale $j\in\{1, ..., M\}$. $M$ is the number of different scales at which a prediction is generated by the network (i.e. $M=6$ in \texttt{ToolNetH}). $\mathbf{\hat{y}}^{(\bar{s})}(\mathbf{z}, \boldsymbol{\theta})$ represents the averaged probabilistic prediction across all scales (i.e. the output of the $1^{2}$ convolution next to all the upsampled predictions in \texttt{ToolNetH}). It is worth noting that by learning the weighting parameters $w_j$ (initialized equally for all the scales) for summing predictions coming from multiple scales, we are learning the contribution from each scale to the final loss. $\bar{\lambda}$ and $\lambda_j$ are hyper-parameters that weight the contribution of the losses at different scales (set to $1$ in our implementation).  


\subsection{Network training details}

Once the network architecture is fixed, the most relevant hyperparameter to tune in a convolutional network trained with \textit{Stochastic Gradient Descent (SGD)} is the learning rate (LR). In our implementation, the learning rate of the network is adjusted following a Cyclical Learning Rate (CLR) policy~\cite{Smith2015}. This policy has shown to drive convolutional networks toward convergence with a higher accuracy in fewer number of optimization steps compared to fixed or exponential policies~\cite{Smith2015}. 

The mechanism of CLR makes the LR oscillate between two boundaries throughout training iterations. There are several waveforms capable of making the aforementioned oscillation. However, all of them have shown comparable results in practice~\cite{Smith2015}. We chose a triangular waveform for the sake of simplicity. 

The triangular CLR has three hyperparameters, minimum and maximum bounds (called \textit{base\_lr} and \textit{max\_lr} respectively) and the number of iterations (called $stepsize$) that it takes from the minimum to the maximum LR. In our implementation, we set $stepsize = 2 \times epoch$, which has been shown to be a convenient value in practice~\cite{Smith2015}. We set the bounds experimentally by running $stepsize$ iterations with several reasonable ranges (different orders of magnitude) and choose the best combination during validation (we split our datasets into three subsets: \textit{training}, \textit{validation} and \textit{testing}). CLR also allows us to fix the number of training iterations. We set it to $6 \times stepsize$ as this has been shown to work well in practice~\cite{Smith2015}.


Momentum and weight decay are set to $0.99$ and $0.0005$ respectively. We use a batch size of $1$. Dropout is used with a probability of $0.5$ only in the smallest scale layer (as can be seen in \cref{fig:cnn_architecture}). Our network architecture is implemented within the Caffe deep learning framework \cite{Jia2014}.

\section{Experiments and Results}

\subsection{Datasets}

To show the robustness and generalization ability of the proposed framework for robotic instrument segmentation, a dataset with different robotic surgical scenarios and instruments has been used to validate the proposed architectures. This dataset consists of training and testing data for \textit{ex vivo} robotic surgery with different articulated instruments.

\subsubsection{Da Vinci Robotic (DVR) dataset \cite{MICCAIchallenge2015,Pakhomov2017}}
The training set contains four \textit{ex vivo} 45-second videos. The first training video features two da Vinci Large Needle Drivers (LNDs) and the rest just one (coming from the right side of the image). The testing set consists of four 15-second and two 60-second videos. The first test video features two LNDs, the fifth and sixth contain two instruments, an LND and a pair of curved scissors (not present in the training set). The rest of test videos feature just one LND (coming from the right side of the image). Articulated motions are present in all the videos. The ground truth masks are automatically generated using joint encoder information and forward kinematics. Hand-eye calibration errors are manually corrected. All the videos have been recorded with the Da Vinci Research Kit (dVRK) Open Source Platform \cite{Kazanzidesf2014}. The frames have a resolution of 720$\times$576 and the videos run at 25 frames per second.


\subsection{Baseline method and evaluation protocol}

As a baseline architecture for comparison, we employ the fine-grained version of the state-of-the-art Fully Convolutional Network (\texttt{FCN-8s}). This network is chosen as a natural baseline for comparison as it represents the state-of-the-art convolutional architecture for robotic surgical tool segmentation \cite{Peraza2016}.

The frames are randomly chosen during training to present the networks with varying input data. As we are mostly interested in comparing the proposed architecture to the baseline method, rather than achieving the highest scores, no data augmentation is performed. For analogous reasons, transfer learning is also neglected. We tune CLR boundaries for all the architectures so that the results are comparable (using the fourth video of the \texttt{DVR} train dataset for validation). In our experiments, the CLR bounds for the \texttt{FCN-8s} network are set to $[1e-10, 1e-8]$. \texttt{ToolNetMS} and \texttt{ToolNetH} are both set to $[1e-7, 1e-5]$. 

The quantitative metrics of choice to evaluate the predicted segmentations are mean Intersection over Union (mean IoU) and mean Dice Similarity Coefficient (mean DSC):
\begin{align}
\overline{\textrm{IoU}}(\mathbf{\hat{y}}, \mathbf{y}) &= \frac{1}{K} \sum_{k=1}^{K} \frac{TP_k}{TP_k + FP_k + FN_k} \\
\overline{\textrm{DSC}}(\mathbf{\hat{y}}, \mathbf{y}) &= \frac{1}{K} \sum_{k=1}^{K} \frac{2TP_k}{2TP_k + FP_k + FN_k}
\end{align}
\noindent where $K = 2$ ($k = 0$ background, $k = 1$ foreground) and $TP_k, FP_k, FN_k$ represent true positives, false positives and false negatives for class $k$ respectively.

All networks were trained and tested (including inference times) on a computer with an Intel Xeon E5-1650 3.50GHz CPU and a NVIDIA GeForce GTX Titan X GPU. The inference time was calculated including data transfers from CPU to GPU and back and averaged across 500 inferences.

\subsection{Quantitative \& qualitative comparison to state-of-the-art}
\begin{table*}
	\centering
	\caption{Quantitative results for segmentation of non-rigid robotic instruments in testing set videos. IoU stands for intersection over union and DSC for Dice Similarity Coefficient. The mean is performed over classes and the results presented are averaged across testing frames.} 
	\begin{tabular}{lccccc}
		\hline
		\multicolumn{1}{c}{\bfseries Network} & 
		\multicolumn{1}{c}{\bfseries ~~No. of parameters~~} &
		\multicolumn{1}{c}{\bfseries ~~Inference time (ms/fps)~~} &
		\multicolumn{1}{c}{\bfseries ~~Balanced Accuracy (fg.)~~} & 
		\multicolumn{1}{c}{\bfseries ~~Mean IoU~~} & 
		\multicolumn{1}{c}{\bfseries ~~Mean DSC~~} \\
		\hline
		\texttt{FCN-8s}     &  $\approx$~134M  &  123.5~/ 8.1           &  77.2\%           &  70.9\%           &  78.8\% \\
		\texttt{ToolNetMS}  &  $\approx$~7.3M  &  \textbf{23.2 / 43.1}  &  78.6\%           &  72.5\%           &  80.4\% \\
		\texttt{ToolNetH}   &  $\approx$~7.4M  &  34.6 / 28.9           &  \textbf{81.0\%}  &  \textbf{74.4\%}  &  \textbf{82.2\%} \\ 
		\hline
	\end{tabular}
	\vspace{0.2cm}
	\label{tab:quantitative_results}
\end{table*}

\begin{figure*}
	\centering
	\includegraphics[width=.91\textwidth]{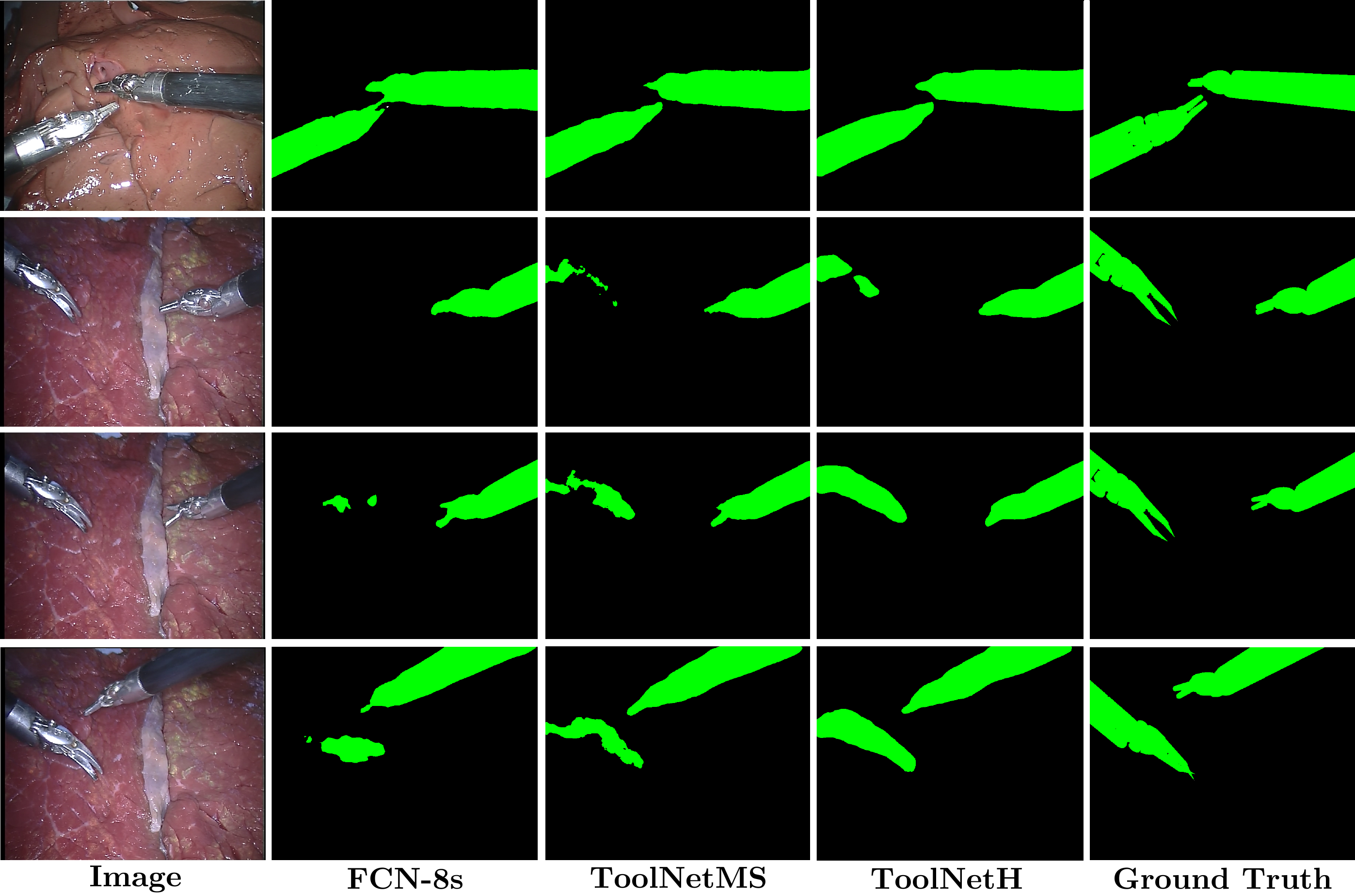}
	\caption{Qualitative comparison of our proposed convolutional architectures versus the state-of-the-art \texttt{FCN-8s} network. Note that the robotic instrument in the left side of the frame is not present in the \texttt{DVR} training set.}
	\label{fig:qualitative_results}
\end{figure*}

Comparing the results of our first proposed architecture \texttt{ToolNetMS} with the state-of-the-art \texttt{FCN-8s} in the \texttt{DVR} testing set, we observed an improvement of $1.4$, $1.6$ and $1.6$ percentage points in balanced accuracy (foreground), mean intersection over union (IoU) and mean DSC respectively (see \cref{tab:quantitative_results}). Furthermore, the model size is reduced by more than an order of magnitude compared to the \texttt{FCN-8s}, from approximately 134M to 7.3M parameters. This has a significant impact on inference speed, which is reduced from $123.5$ms for the \texttt{FCN-8s} to $23.2$ms for the \texttt{ToolNetMS}, becoming a viable real-time instrument-tissue segmentation method for robotic surgery with the da Vinci platform.

The results of our holistically-nested method \texttt{ToolNetH} show an improvement over \texttt{FCN-8s} of $3.8$, $3.5$ and $3.4$ percentage points in balanced accuracy (foreground), mean IoU and mean DSC respectively (see \cref{tab:quantitative_results}). The inference speed is also real-time, approximately $29$fps. The number of parameters is 7.4M, comparable to the 7.3M of \texttt{ToolNetMS}.

The qualitative results in \cref{fig:qualitative_results} show how our proposed architecture respects the borders of the surgical tools more accurately than the competing state-of-the-art architecture and provides a segmentation mask that shows enhanced spatial consistency.

\section{DISCUSSION AND CONCLUSIONS}

We have proposed novel multi-scale and holistically-nested CNN architectures trained with Dice loss function for real-time segmentation of robotic surgical tools, at a frame rate of approximately $29$Hz. 

Our proposed methods show a competitive advantage over state of the art because they take into account the issues faced by prior methods (8x upsampling of the prediction in \texttt{FCN-8s} and lack of real-time), the details of the problem (need for segmentations that respect the edges of the robotic tools, so that shape can be accurately used for control purposes) and data (unbalancing of pixels between foreground and background). The encouraging results observed with test data suggest that fusing convolutional feature maps from all the intermediate layers of the CNN helps the segmentation of robotic surgical tools, and imposing deep supervision at all the intermediate layers further improves the segmentation performance. 

We have shown that our trained model is robust and generalizes well to previously unseen robotic instruments and video recording conditions. The architecture is general and we believe it will be of research interest for tasks similar to real-time surgical instrument segmentation such as tool detection or surgical phase recognition.  Moreover, the proposed CNN architecture has up to $15$ times fewer parameters than the widely-used FCNs, which is key to achieving real-time performance. This significant reduction in model size also enables easy distribution of our model to end-users and deployment to embedded devices.

In the future, rather than dramatically reducing the number of parameters, we will investigate how to optimise the architecture to achieve higher performance while maintaining the given computational budget. Data augmentation and transfer learning, possibly from deeper architectures, are also a matter for further study. Adversarial training is also an element of further investigation, as it has shown positive results in general purpose semantic labeling applications and it could potentially help to further increase the segmentation accuracy of robotic surgical instruments.

Localization of the segmented instruments in 3D arises as interesting future work. As stereo endoscopes become more widespread, both images could be segmented, guiding the matching of instrument pixels between them. Provided that the cameras are calibrated, these matches allow for dense geometry based triangulation of the surgical tool.

\section*{ACKNOWLEDGMENTS}

This work was supported by Wellcome Trust [WT101957, 203145Z/16/Z], EPSRC (NS/A000027/1, EP/H046410/1, EP/J020990/1, EP/K005278, NS/A000050/1), NIHR BRC UCLH/UCL High Impact Initiative and a UCL EPSRC CDT Scholarship Award (EP/L016478/1). The authors would like to thank NVIDIA Corporation for the donated GeForce GTX TITAN X GPU and all the members of the GIFT-Surg project for the always useful discussions.

\bibliographystyle{IEEEtran}
\bibliography{IEEEabrv,library}

\end{document}